\documentclass{article}

\usepackage[preprint]{neurips_2023}

\usepackage[utf8]{inputenc} 
\usepackage[T1]{fontenc}    
\usepackage{hyperref}       
\usepackage{url}            
\usepackage{booktabs}       
\usepackage{amsfonts}       
\usepackage{nicefrac}       
\usepackage{microtype}      
\usepackage{xcolor}         

\usepackage{amsmath}
\usepackage{graphicx}
\usepackage{changepage}
\usepackage{tabu}
\usepackage{rotating}

\usepackage{textcmds}
\usepackage{subcaption}
\usepackage{multirow}

\usepackage{tikz}
\usepackage{amsmath,bm,times}
\usetikzlibrary{calc}

\usepackage{sistyle}
\SIthousandsep{,}

\title{Environmental-Impact Based Multi-Agent Reinforcement Learning}

\author{
	Farinaz Alamiyan-Harandi \\
	Department of Electrical \& Computer Engineering \\
	Isfahan University of Technology,
	Isfahan, Iran \\
	\texttt{farinaz.alamiyan@gmail.com} \\ 
	\And
	Pouria Ramazi \\
	Department of Mathematics \& Statistics \\ Brock University, St. Catharines, Canada\\  
	\texttt{prramazi@brocku.ca} \\
}

\begin{document}
	
\pgfdeclarelayer{background}
\pgfdeclarelayer{foreground}
\pgfsetlayers{background,main,foreground}

\tikzstyle{states}=[dotted, black, fill=green!40, text width=1.2em, text centered, minimum height=1.7em, thick]

\tikzstyle{layers} = [draw, thick, text centered, text width=2.5em, fill=white!20, 
minimum height=6em]
\def\blockdist{0.5}
\def\edgedist{1}

\maketitle

\begin{abstract}
  To promote cooperation and strengthen the individual impact on the collective outcome in social dilemmas, we propose the Environmental-impact Multi-Agent Reinforcement Learning (EMuReL) method where each agent estimates the ``environmental impact'' of every other agent, that is, the difference in the current environment state compared to the hypothetical environment in the absence of that other agent. Inspired by the Inequity Aversion model, the agent then compares its own reward with those of its fellows multiplied by their environmental impacts. If its reward exceeds the scaled reward of one of its fellows, the agent takes “social responsibility” toward that fellow by reducing its own reward. Therefore, the less influential an agent is in reaching the current state, the more social responsibility is taken by other agents. Experiments in the Cleanup (resp. Harvest) test environment demonstrate that agents trained based on EMuReL learn to cooperate more effectively and obtain $54\%$ ($39\%$) and $20\%$ ($44\%$) more total rewards while preserving the same cooperation levels compared to when they are trained based on the two state-of-the-art reward reshaping methods inequity aversion and social influence.
\end{abstract}

\section{Introduction}
\label{S.Introduction}
Many real-world problems are selfish multi-agent systems with shared facilities, referred to as the \emph{common pool resource problems} \citep{gardner1990nature}. The agents choose from a number of available actions to use the shared facility and accordingly earn \qq{rewards} from the environment. Using the common resource to maximize its own reward, each agent's goal will be in potential conflict with that of its fellows. Examples include water resource management \citep{pretorius2020game}, real-time traffic flow control \citep{chu2020multi}, coordination of autonomous vehicles \citep{sallab2017deep}, and multi-player video games \citep{kempka2016vizdoom}.
In the autonomous vehicle problem, streets, intersections, and highways are shared common resources. Each vehicle plans a path to reach its given destination and is rewarded based on how fast it gets there. Ideally, each vehicle should choose the shortest path, but this may lead to congestion. So to maximize their individual rewards, they need to cooperate. Such situations raise a type of Sequential Social Dilemma (SSD) \citep{leibo2017multi} where the most rewarding strategy for each agent (in the short-term) is to \qq{defect} and exploit the common resource as much as possible, resulting in a mutual defection among all agents; however, this results in the depletion of the common resource and each agent earns (in the long-term) much less than if they would have all \qq{cooperated} and used the common resource up to a certain limit. In SSDs, cooperative or defective behaviors exist not only as atomic actions but they are temporally extended over each agent's decision-making \qq{policy}, that is, a strategy that determines what action to choose at each environment state. The question then is how to design the agents' policies to have cooperative agents that maximize the collective reward, i.e., to maximize total reward while minimizing unfairly low rewards earned by cooperators who facilitate the high rewards earned by the \qq{free-riders} who defect. 

The many often stochastic factors impacting the environment and their unknown underlying mechanisms challenge the use of traditional modeling techniques in solving SSDs. For example, in the autonomous vehicle problem, the number of vehicles in each lane is variable, and great uncertainty is involved in the drivers’ behavior. Moreover, the observation and action spaces of each agent are too large to search through exhaustively or near exhaustively to find the optimal policies. Even if the optimal policies are found, they are not generalizable – a new environment would need new policies. 

To tackle these modeling complexities, in addition to model-based structures, Multi-Agent Reinforcement Learning (MARL) proposes a model-free structure where intelligent agents require no prior knowledge about the environment and \qq{learn} the optimal policies directly from interacting with the environment. During the interactions, each agent observes the environment state and executes one of the available actions based on its policy. Consequently, the environment transfers to a new state and provides the agent an \textit{extrinsic reward} indicating the immediate desirability of the environment state. The agent utilizes these interactions to estimate a \textit{state value function}, that assigns to each state the long-term expected reward that the agent obtains when starting from that state and following its policy. According to this value function, the agent improves its policy to choose those actions that result in the highest-value states \citep{sutton1998introduction}. The learned policies may be generalized to or sometimes readily used in unseen environments of the problem. 

However, learning policies in SSDs by using only extrinsic rewards has limited performance, because each extrinsic reward is agent (rather than collective) specific and typically selfishly maximized by exploiting other agents, which can lead to mutual defection. This is because a cooperative equilibrium either does not exist or is difficult to find\citep{hernandez2018multiagent,heemskerk2020social}. One way to mitigate social dilemmas is to reshape the rewards so that defecting behaviours are no longer an equilibrium. To this end, researchers have enriched extrinsic rewards by the so-called \textit{intrinsic rewards} \citep{barto2005intrinsic} that capture aspects of the agents' behavior that are not necessarily encoded by the extrinsic reward. For each agent $k$ at time step $t$, this results in its \textit{reshaped reward} $r_{t}^{k}$, defined as a linear combination of its extrinsic reward $e_{t}^{k}$ and intrinsic reward $i_{t}^{k}$:  
\begin{equation}\label{equ.general_reward}
\begin{aligned}
r_{t}^{k}=\alpha e_{t}^{k} +\beta i_{t}^{k} .
\end{aligned}
\end{equation}
for constant scalers $\alpha$ and $\beta$.

In the Inequity Aversion (IA) model\citep{hughes2018inequity}, the intrinsic reward is calculated based on envy and guilt. Each agent compares its own and fellows' extrinsic rewards to detect inequities and balances its selfish desire for earning rewards by keeping the differences as small as possible, see \eqref{equ.Inequity_Aversion_intrinsic_reward}.
Here, the feeling of guilt can be interpreted as a feeling of social responsibility; that is, the agent wants to revise its policy to reduce the reward inequity of itself compared to change their \qq{worse-performing} others. However, the IA model measures the agents' performance only by their obtained rewards and ignores their roles in establishing the current environment state that resulted in those rewards. This may lead to equal treatment of defectors and cooperators that earn the same reward, despite their different contributions to the obtained rewards. Namely, rather than only those who earned less than itself, each agent should \qq{feel responsible} for also those who earned more but did not contribute to reaching the current rewarding environment state. Even among equally less-earned others, a higher social responsibility should be dedicated to those who also contributed less to reach the current state. Similarly, compared to higher-earning other agents, the agent feels that it should work \qq{harder} (by finding a better policy) to earn as much as they do, and this feeling should be stronger toward those who played a more effective role in changing the state.

To take into effect the agents' roles in reaching the environment states, we propose an \emph{environmetal impact} of each agent defined by the difference between the current local state and the hypothetical one where that agent would have been absent in the previous state. In the autonomous vehicle example, a vehicle's impact can be the difference in the current congestion in the presence and absence of that vehicle. When performing the comparisons in reshaping the reward function, each agent $k$ computes the impact $d^{k,j}_t$ of every other agent $j$ and scales that agent's reward $e^j_t$ by its impact $d^{k,j}_t$. We demonstrate the effectiveness of the proposed impact criterion through experiments in the Cleanup and Harvest environments \citep{hughes2018inequity,  perolat2017multi, SSDOpenSource}. The results demonstrate that agents trained by the EMuReL method can learn to cooperate more effectively compared with the IA and \emph{Social Influence} (SI) \citep{jaques2019social} methods  that is, they earn a higher collective reward and have a slightly higher cooperation level. 
 
The rest of this paper is organized as follows: The MARL setting in an SSD environment and the basic formulation of the proposed Environmental Impact based Multi-Agent Reinforcement Learning (EMuReL) approach are explained in Section \ref{S.MARL_in_SSDs_Environmental_Impact}. Section \ref{S.Experiments} includes the experiments and comparison results. The discussion is given in Section \ref{S.Discussion}.
\section{Related Works}
\label{S.Related_Works}
Insights into the nature of human social behavior have led to the introduction of two main categories of intrinsic rewards. 
The first focuses on self-preferencing attributes, including empowerment \citep{klyubin2005empowerment}, social influence \citep{jaques2019social}, and curiosity \citep{pathak2017curiosity,burda2018large}. The second is motivated by social welfare emotions, including envy and guilt \citep{hughes2018inequity}, and empathy \citep{salehi2019empathetic}. 
For example, the IA model \citep{hughes2018inequity} is based on envy and guilt caused by inequities in the extrinsic rewards.
A \textit{disadvantageous inequity} occurs when agent $k$ experiences \qq{envy} by earning a lower reward than another agent $j$, that is, if $e^{j}_{t} - e^{k}_{t}>0$. Oppositely, an \textit{advantageous inequity} happens when the agent experiences \qq{guilt} by earning a higher reward than another agent $j$, that is, if $e^{k}_{t} - e^{j}_{t}>0$. To incorporate its feelings of envy and guilt, agent $k$ averages these comparisons over all other agents, resulting in the following intrinsic reward:
\begin{equation}\label{equ.Inequity_Aversion_intrinsic_reward}
\begin{aligned}
i_{t}^{k} = -\frac{\alpha_{k}}{N-1}\sum_{j\neq k} \max(e^{j}_{t} - e^{k}_{t}, 0) - \frac{\beta_{k}}{N-1}\sum_{j\neq k} \max(e^{k}_{t} - e^{j}_{t}, 0) ,
\end{aligned}
\end{equation}
where $N$ is the total number of agents and parameters $\alpha_{k}$,$\beta_{k}\in\mathbb{R}$ control the agent’s aversion to disadvantageous and advantageous inequities, respectively. Regardless of whether it earns more or less than others, the agent decreases its obtained reward. 

Other types of environment \qq{impacts}, such as auxiliary reward functions \citep{turner2020conservative} and measure of deviation for some baseline state \citep{krakovna2018penalizing}, have been defined in the literature to avoid undesired irreversible changes to the environment.

Two influence-based methods are proposed by \citet{wang2019influence} to solve the multi-agent exploration problem. These methods exploit the interactions among agents and measure the amount of one agent’s
influence on the other agents’ exploration processes by using mutual information between agents’ trajectories.

\section{Background: MARL and
Markov games}
\label{S.background}
A Markov game \citep{littman1994markov} is a standard framework for modeling MARL problems such as SSDs. It is defined on a state space $\mathcal{S}$, a collection of $N$ agents' action sets $\mathcal{A}=\{ \mathcal{A}^1, ..., \mathcal{A}^N \}$, and a state \textit{transition distribution} $\mathcal{T}$ preserving the Markov property, that is, the next state of the environment is independent of the past states and depends only on the current environment state and the agents' applied actions \citep{markov1954theory,van1981stochastic}. The global state of the environment at time step $t$ is given by $s_{t}\in \mathcal{S}$. Each agent $k$ observes the environment to some extent as a local state $s^{k}_{t}$ and selects an action $a^{k}_{t}\in \mathcal{A}^k$ based on its \textit{policy} $\pi^{k}$, that is, the probability distribution of selecting each of the available actions. A \emph{joint action} $\bm{a}_{t}$ is the stacking of all agents’ actions $[a^{1}_{t},...,a^{k}_{t},...,a^{N}_{t}]$. Applying the joint action $\bm{a}_{t}$ at the global state $s_{t}$ causes a transition in the environment state according to the transition distribution  $\mathcal{T}(s_{t+1}|s_{t},\bm{a}_{t})$. Each agent $k$ then receives a reward $r^{k}_{t+1}$, that is, originally the agents' extrinsic reward $e^{k}_{t+1}$ or can be reshaped by utilizing some intrinsic reward $i^{k}_{t+1}$. Using a Reinforcement Leaning (RL) algorithm, the agent evaluates its policy by calculating its associated \textit{value function} $V^{\pi^{k}}(s^{k}_{t})=\mathbb{E}[R^{k}_{t}|s_t^k,\pi^k]$ over every state $s^k_t$, where $R^{k}_{t}=\sum_{i=0}^{\infty}\gamma^{i}r^{k}_{t+i+1}$ is the cumulative discounted future rewards starting from the current local state $s_t^k$ following policy $\pi^{k}$ where $\gamma\in[0,1]$ is the \textit{discount factor}. Then agent $k$ learns to improve its policy such that in each state $s^{k}_{t}$, the agent chooses the action that transfers the environment to the next state $s^{k}_{t+1}$ where $V^{\pi^{k}}$ is maximized.
\section{The EMuReL approach}
\label{S.MARL_in_SSDs_Environmental_Impact}
To define intrinsic rewards, we use the same equation as in (\ref{equ.Inequity_Aversion_intrinsic_reward}) with two main differences. First, as in \citep{hughes2018inequity}, to take into account the temporally distributed, rather than the single-instant reward, we replace the extrinsic reward $e^j_t$ with 
the \textit{temporal smoothed extrinsic rewards} $w^{j}_{t}$ defined by
\begin{equation}\label{equ.Inequity_Aversion_e}
\begin{aligned}
w^{j}_{t} = \gamma \lambda w^{j}_{t-1} + e_{t}^{j}  \;\;\;\;\;\;\;\forall t\geq 1, \; w^j_0 = 0,
\end{aligned}
\end{equation}
where $\lambda\in[0,1]$ is a hyperparameter. Second, when performing the IA comparisons, agent $k$ scales the reward of every agent $j$ by $d^{k,j}_{t}\in[0,1]$, that is, agent $j$'s \textit{environmental impact} in establishing agent $k$'s local state at time $t$. Hence, agent $k$'s intrinsic reward is
\begin{equation}\label{equ.impact_Inequity_Aversion_reward}
\begin{aligned}
i_{t}^{k} = -\frac{\alpha_{k}}{N-1}\sum_{j\neq k} \max(d^{k,j}_{t}w^{j}_{t} - w^{k}_{t}, 0) - \frac{\beta_{k}}{N-1}\sum_{j\neq k} \max(w^{k}_{t} - d^{k,j}_{t}w^{j}_{t}, 0),
\end{aligned}
\end{equation}
where ${\alpha}_{k},{\beta}_{k}\in\mathbb{R}$. The definition of the impact $d^{k,j}_{t}$ lies in the answer to the following question:
\begin{quote}
	Agent $k$: \qq{How impactful was the presence of agent $j$ in reaching my current local environment state $s^{k}_{t}$?}	
\end{quote}
So agent $k$ needs to compare its current local state $s^k_t$ in the presence and absence of agent $j$. This requires estimating the typically large size and detailed state $s^k_t$ in the absence of agent $j$ with a reasonable accuracy, which may not be practical. Moreover, $s^k_t$ is often an image with many features that may not have been caused by the agents' actions. Therefore, rather than $s^k_t$, we use a reduced-dimension \emph{feature encoding function} $\phi(s^k_t)$ that encodes only those features of the environment state influenced by the agents’ actions \citep{pathak2017curiosity}.

Now, inspired by dropouts technique used in neural network regularization \citep{srivastava2014dropout} to be able to extract the features of the local state $s^k_t$ in the absence of agent $j$, we need to estimate $\phi(s^k_t)$ when agent $j$ is omitted. This results in the \emph{estimated feature encoding function} $\hat\phi$ that instead of $s^k_t$, takes the previous joint action $\bm{a}_{t-1}$ and features $\phi(s^k_{t-1})$ as its input and estimates $\phi(s^k_t)$ as its output, i.e.,
\begin{equation}
\begin{aligned}
\hat{\phi}(\phi(s^k_{t-1}),\bm{a}_{t-1}) \approx \phi(s^k_{t}) .
\end{aligned}
\end{equation}
Then by omitting $a^j_{t-1}$, resulting in the reduced joint action denoted by $\bm{a}^{-j}_{t-1}$, we obtain the estimated features at state $s^k_{t}$ in the absence of agent $j$ (Figure \ref{fig.EMuReL_elimination}). 
Taking the norm of the difference of the two estimated features results in the impact of agent $j$ in view of agent $k$:
\begin{equation}\label{equ.EMuReL_d}
\begin{aligned}
d^{k,j}_t = \frac{1}{2} \left \| \hat{\phi}(\phi(s^k_{t-1}),\bm{a}_{t-1}) - \hat{\phi}^{-j}(\phi(s^k_{t-1}),\bm{a}^{-j}_{t-1}))\right \| _{2}^{2},
\end{aligned}
\end{equation}
where $\|\cdot\|_2$ is the Euclidean norm and $\hat{\phi}^{-j}$ is the same as $\hat{\phi}$ but when agent $j$ is eliminated. These disparities are scaled using unity-based normalization to bring all values into the range $[0,1]$. We emphasize that the elimination of an agent is different from when the agent is present but performs no operation (NOOP).

To learn the feature encoding function $\phi$ and its estimation $\hat\phi$, inspired by the neural network structure of Social Curiosity Module (SCM) \citep{heemskerk2020social} for multi-agent systems, we extend the Intrinsic Curiosity Module (ICM) \citep{pathak2017curiosity} in the single-agent setup to our multi-agent RL system. We propose the Extended ICM (EICM) by embedding two associated predictors inside the architecture of each agent $k$ (Figure \ref{fig.EICM}). 

First is the \textit{forward model} representing $\hat{\phi}$ by a neural network function $f$ with parameters $\theta_F$ that predicts the encoded state $\phi(s^k_t)$ based the previous joint action $\bm{a}_{t-1}$ and encoded state $\phi(s^k_{t-1})$. Nevertheless, the state $s^k_t$ is the result of applying the agents' actions $\bm{a}_{t-1}$ on the previous global state $s_{t-1}$, not local state $s^k_{t-1}$. Hence, the forward model needs the global state $s_{t-1}$ or an estimation of it. To this end, we exploit the Model of Other Agents (MOA) \citep{jaques2019social} that is a neural network embedded in each agent's architecture to predict the other agents' actions based on their previous actions and the previous state, i.e., $P(\bm{a}_t \mid s^k_{t-1},\bm{a}_{t-1})$. Via its internal LSTM  state $u_{t-1}^k$, MOA implicitly models the state transition function $\mathcal{T}(s_{t}|s_{t-1},\bm{a}_{t-1})$ to estimate the global state $s_{t}$ (the green nodes in the gray rectangle in Figure \ref{fig.EICM}) or $s_{t-1}$ when the time is shifted negatively by one unit, i.e., $\mathcal{T}(s_{t-1}|s_{t-2},\bm{a}_{t-2})$. By passing this internal LSTM state to the forward model, we implicitly provide an estimate of the global state $s_{t-1}$. Therefore, the function $f$ can be written as 
\begin{equation}\label{equ.EMuReL_functions1}
\begin{aligned}
\hat{\phi}(\phi(s^{k}_{t-1}),\bm{a}_{t-1}) &=f(\phi(s^{k}_{t-1}), u^{k}_{t-1} , \bm{a}_{t-1}; \theta_{F} ) . 
\end{aligned}
\end{equation}
\begin{center}
	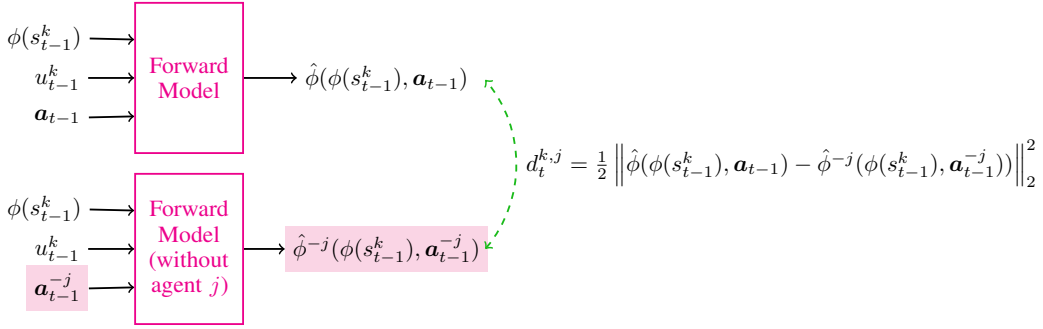
\begin{figure}[!h]
		\centering
		\resizebox{\columnwidth}{!}{
			\begin{tikzpicture} 
			\node (Input11) at (-0.2,1.9) [] {$\phi(s^k_{t-1})$};
			\node (Input12) at (0,1.3) [] {$u^k_{t-1}$};
			\node (Input13) at (0,0.7) [] {$\bm{a}_{t-1}$};
			\node (Input21) at (-0.2,-0.7) [] {$\phi(s^k_{t-1})$};
			\node (Input22) at (0,-1.3) [] {$u^k_{t-1}$};
			\node (Input23) at (0,-1.9) [fill=magenta!20] {$\bm{a}^{-j}_{t-1}$};
			\node (ForwardModel1) at (2,1.3) [layers, text width=4em, rectangle, draw=magenta!100, minimum height=6.5em, text=magenta!100, thick] {Forward Model};
			\node (ForwardModel2) at (2,-1.3) [layers, text width=4em, rectangle, draw=magenta!100, minimum height=6.5em, text=magenta!100, thick] {Forward Model (without agent $j$)};
			\node (Output1) at (5,1.3) [] {$\hat{\phi}(\phi(s^k_{t-1}),\bm{a}_{t-1})$};
			\node (Output2) at (5,-1.3) [fill=magenta!20] {$\hat{\phi}^{-j}(\phi(s^k_{t-1}),\bm{a}^{-j}_{t-1})$};
			\draw [->, thick] (Input11.east) -- (ForwardModel1.145);  
			\draw [->, thick] (Input12.east) -- (ForwardModel1.180); 
			\draw [->, thick] (Input13.east) -- (ForwardModel1.215);
			\draw [->, thick] (ForwardModel1.east) -- (Output1.west); 
			\draw [->, thick] (Input21.east) -- (ForwardModel2.145);  
			\draw [->, thick] (Input22.east) -- (ForwardModel2.180); 
			\draw [->, thick] (Input23.east) -- (ForwardModel2.215); 
			\draw [->, thick] (ForwardModel2.east) -- (Output2.west); 
			\coordinate (Origin) at (4.5,-1.3);
			\draw [<->, draw=black!30!green!90, thick, dashed] ($(Origin)+(2,0)$) 
			arc (-40:40:2);
			\node at (11,0) {$d^{k,j}_t = \frac{1}{2} \left \| \hat{\phi}(\phi(s^k_{t-1}),\bm{a}_{t-1}) - \hat{\phi}^{-j}(\phi(s^k_{t-1}),\bm{a}^{-j}_{t-1}))\right \| _{2}^{2}$};
			\end{tikzpicture}}
		\caption{\textbf{The elimination process of computing impacts by agent $k$.} In the iterative procedure of this process, each time, an agent is removed from the neural network computations of the forward model by setting the corresponding network weights of that agent in the input to zero. The disparity between the outputs of the forward model in the presence and absence of agent $j$ is employed as a measure of agent $j$'s impact in predicting the features of the current local state from the agent $k$'s viewpoint.} 
		\label{fig.EMuReL_elimination}
	\end{figure}
\end{center}
During the training of the forward model, parameters $\theta_{F}$ and $\theta_{\phi}$ are learned to minimize $L_F$ defined as the discrepancy between the predicted and actual encoded local states: 
\begin{equation}\label{equ.ICM_reward}
\begin{aligned}
L_{F}\left(\phi(s^{k}_{t}), \hat{\phi}\left(\phi(s^{k}_{t-1}),\bm{a}_{t-1}\right)\right) = \frac{1}{2} \left \| \hat{\phi}(\phi(s^{k}_{t-1}),\bm{a}_{t-1}) - \phi(s^{k}_{t}) \right \| _{2}^{2} .
\end{aligned}
\end{equation}
Second, to lead the forward model to extract those features that are more relevant to the agents' actions, the actions are estimated using the so called \textit{inverse model} and parameters $\theta_\phi$ are tuned by minimizing the loss of the actions estimation. The inverse model is a neural network function $g$ with parameters $\theta_I$ that predicts the applied joint action $\bm{a}_{t-1}$ given $\phi(s^k_{t-1})$ and $\phi(s^k_{t})$ as follows:
\begin{equation}\label{equ.EMuReL_functions2}
\begin{aligned}
\hat{\bm{a}}_{t-1} &=g(\phi(s^{k}_{t-1}), \phi(s^{k}_{t}), u^{k}_{t-1}; \theta_{I} ) .
\end{aligned}
\end{equation} 

Then parameters $\theta_{I},\theta_{\phi}$ are learned to minimize $L_{I}(\hat{\bm{a}}_{t-1}, \bm{a}_{t-1})$ where $L_{I}$ is a cross entropy over the predicted and actual actions of the agents: 
\begin{equation}\label{equ.ICM_LI}
\begin{aligned}
L_{I}(\hat{\bm{a}}_{t-1}, \bm{a}_{t-1}) = -\sum_{j=1}^{N} a^j_{t-1}\log(\hat{a}^j_{t-1}) .
\end{aligned}
\end{equation}
\section{Experiments}
\label{S.Experiments}
In this section, two SSD games and the result of applying the proposed EMuReL as well as its comparison with some baseline methods are presented. 
\subsection{Experimental setup}
\label{S.Experiments.Experimental_Setup}
Multi-agent sequential social dilemmas are divided into two main categories: 1) public goods dilemmas, where providing a shared resource requires each agent to pay a personal cost, and 2) commons dilemmas, where defecting causes depletion of a shared resource \citep{kollock1998social}. The Cleanup and Harvest games are examples of these dilemmas, respectively (See Figure \ref{fig.1_cleanup_harvest}). In the Cleanup game, there are two geographically separated areas on a two-dimensional grid environment: an apple field and a river. A group of agents moves inside the field and collects apples to obtain rewards. For producing apples, the agents should use their cleaning beam and clean up some of the waste that appears in a river over time. A higher waste level means a lower apple reproducing-rate in the field. Similar to the Cleanup game, the Harvest game includes a discrete grid area as an apple field with a number of agents that collect apples to earn rewards. However, the growth rate of new apples is determined by the apple density in each area of the field. The higher the apple density in an area, the faster new apples grow in that area. When all apples in an area are harvested, none will ever grow back. So agents should choose an appropriate harvesting rate to both maximize their reward and keep an ongoing apple reproduction rate. In both games, the extrinsic reward function is the same: $+1$ is the reward for collecting each apple. The agents are equipped with a punishment beam as an action to fire others with the cost of -1. An agent hit by this beam will lose $50$ rewards. The Cleanup and Harvest environments were developed by Vinitsky et.al \citep{SSDOpenSource} as open-source Python codes.

As illustrated in the Schelling diagrams by \citet{hughes2018inequity}, these games are SSDs. In the Cleanup game, if an agent defects by staying in the apple field longer without cleaning up the river, it can obtain a higher reward. However, too many agents defecting makes the apple field depleted, resulting in lower future rewards. So increasing the number of cooperative agents improves the overall long-term reward for a single cooperator. In the Harvest game, if an agent defects and collects all nearby apples quickly, it can receive a higher reward. However, continuing this manner causes an unproductive empty field in the long term.

To learn the policy and value function, a decentralized setting of MARL with no communication access is used. Each agent uses a neural network for each of the policy and value function based on the structure of the \textit{SI method} \citep{jaques2019social}, where each network consists of \emph{(i)} a convolutional layer to handle image inputs ($15\times 15$ pixels images), \emph{(ii)} some fully connected layers to mix signals of information between each input dimension and each output, \emph{(iii)} a Long Short Term Memory (LSTM) recurrent layer to create an internal memory, and \emph{(iv)} some linear layers to provide the full range of output values. To learn the parameters of the two neural networks, the policy gradient algorithm Proximal Policy Optimization (PPO) \citep{schulmanwdrk2017ppo} and the Asynchronous Advantage Actor-Critic (A3C) \citep{mnih2016asynchronous} are used. These algorithms take an \textit{actor-critic} structure that optimizes the parameters of neural networks of the policy (actor) and the value function (critic) with respect to the gradient of the actor estimated by the critic using gradient ascent.

Here, the results of four methods were compared: \emph{(i)} the \emph{baseline method} that uses only the extrinsic rewards, \emph{(ii)} the \emph{IA method} that utilizes the reshaped rewards based on the IA intrinsic rewards (\ref{equ.Inequity_Aversion_intrinsic_reward}), \emph{(iii)} the \emph{SI method} \citep{jaques2019social} that enriches the extrinsic rewards with intrinsic rewards derived from social empowerment, having a causal influence on other agents to make them change their policies, and \emph{(iv)} our proposed \emph{EMuReL method} that enriches the IA intrinsic rewards with the environmental impacts. The SCM algorithm with the same parameter setting as in \citet{heemskerk2020social} was adapted to the proposed EMuReL method in the Cleanup and Harvest environments. Two commonly used algorithms in these environments are PPO and A3C. As the baseline and SI methods with the PPO algorithm are reported to achieve higher collective rewards than with the A3C algorithm in the Cleanup environment \citep{jaques2019social,heemskerk2020social}, we chose PPO for this environment. For the Harvest environment, we tested both PPO and A3C algorithms. An ablation system was also used to test the contribution of each key ingredient of the EMuReL Method.  

For the hyperparameters, the default values 
recommended by the authors of the PPO algorithm were used \citep{schulmanwdrk2017ppo}. The training batch size and the PPO minibatch size as two effective hyperparameters were set to $\num{96000}$ and $\num{24000}$, respectively. 
According to the results reported by \citet{hughes2018inequity}, the advantageous-IA agents are more effective in Cleanup game, and having more agents who are averse to inequity facilitates cooperation. Thus in this environment, for the IA and EMuReL methods, we considered $5$ advantageous-IA agents with $\beta$ equal to $0.05$ and setting $\alpha=0$. We conducted $15$ experiments of each method with random seeds and without optimizing the hyperparameters. For the Harvest environment, we evaluated both advantageous-IA and disadvantageous-IA agents and conducted $5$ experiments of each method. We consider disadvantageous-IA agents with $\alpha$ equal to $5$ and setting $\beta=0$. Every experiment was performed on a Linux server with $3$ CPUs, a P$100$ Pascal GPU, and $100$G RAM and took between $10$ and $28$ days considering the setting.

\subsection{Experimental results}
\label{S.Experiments.Experimental_Results}
\begin{center}
	\begin{figure}[!h]
		\centering			
		\resizebox{\columnwidth}{!}{
			\extrarowsep=_3pt^3pt			
			\begin{tabu}to\linewidth{cccc}
				\raisebox{-0.5\height}{\includegraphics[width=0.5\columnwidth ]{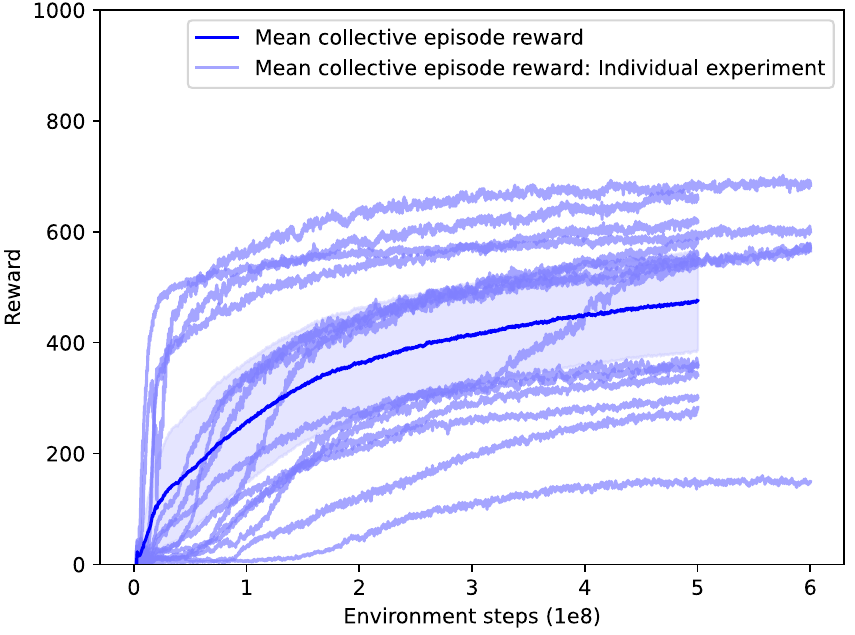}} &
				\raisebox{-0.5\height}{\includegraphics[width=0.5\columnwidth ]{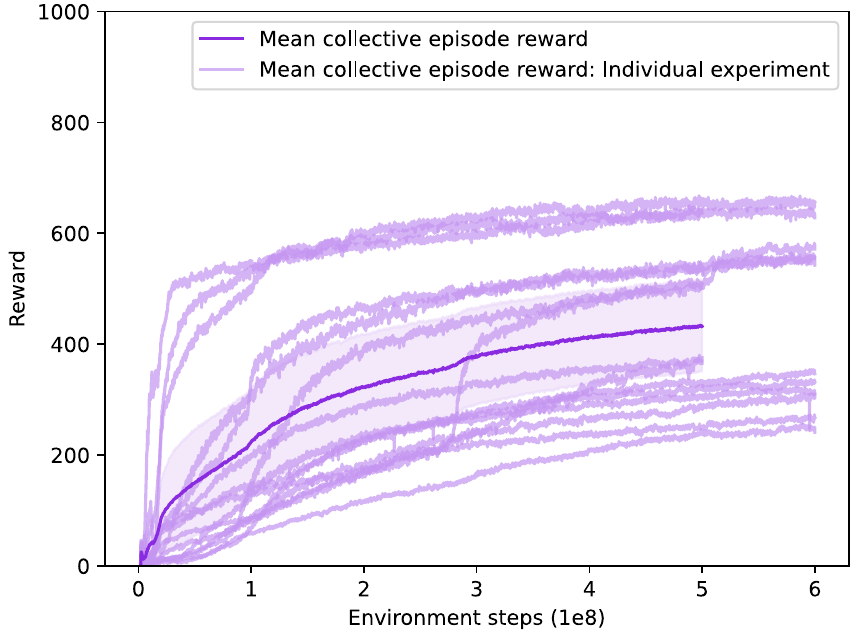}} &  
				\raisebox{-0.5\height}{\includegraphics[width=0.5\columnwidth ]{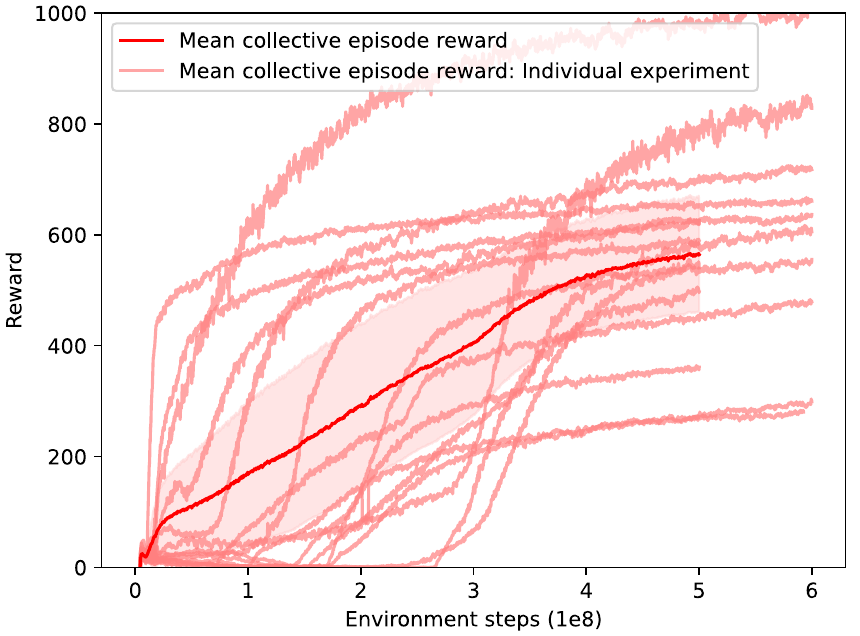}}	& \raisebox{-0.5\height}{\includegraphics[width=0.5\columnwidth ]{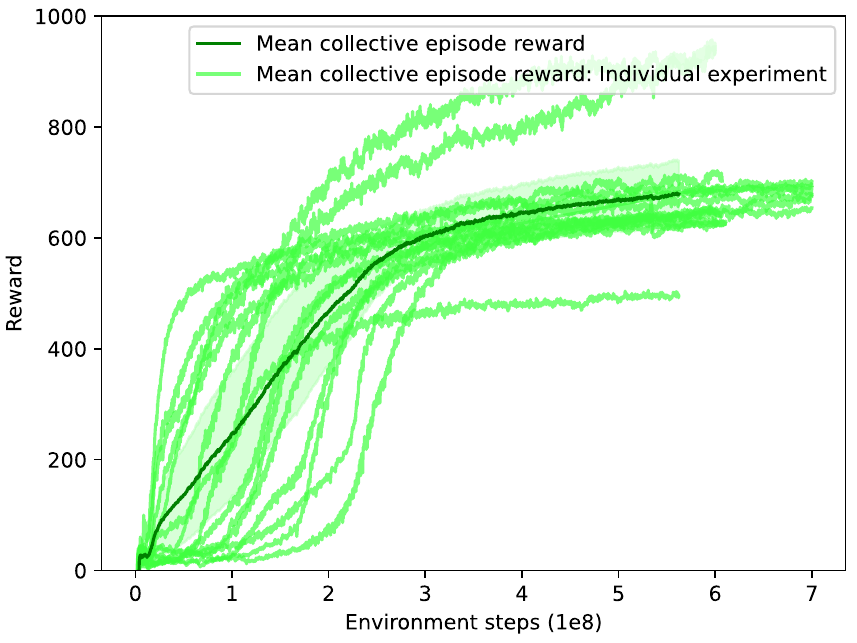}}
				\\
				\LARGE (a) Baseline & 
				\LARGE (b) IA & \LARGE (c) SI	 & \LARGE (d) EMuReL  \\			 
				\multicolumn{2}{c}{\raisebox{-0.5\height}{\includegraphics[width=0.8\columnwidth ]{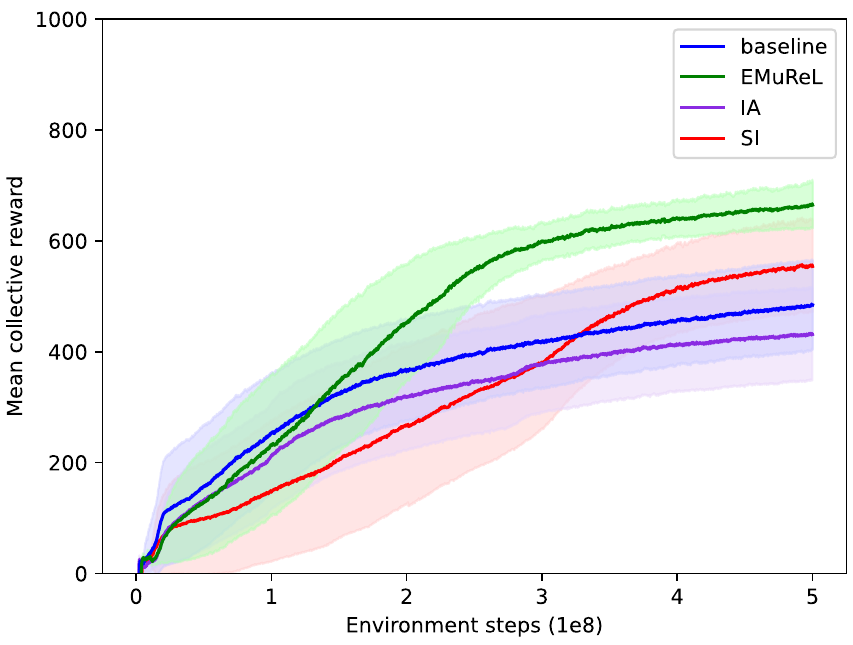}}} &				 
				\multicolumn{2}{c}{\raisebox{-0.5\height}{\includegraphics[width=0.8\columnwidth ]{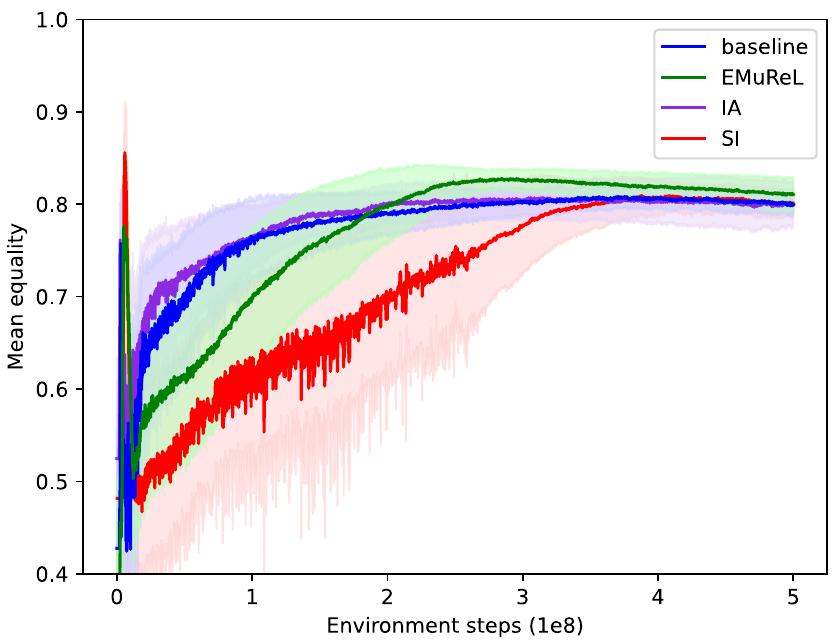}}}
				\\ 
				\multicolumn{2}{c}{\LARGE (e) Reward comparison} & \multicolumn{2}{c}{\LARGE (f) Equality comparison}	 
		\end{tabu}}
		\caption{\textbf{The results for the Cleanup environment.} (a-d) The mean reward obtained by $5$ agents over individual experiments of the baseline, IA, SI, and EMuReL methods using PPO algorithm. Each point of these curves shows the average collective reward over at least $\num{96000}$ environment steps ($96$ episodes of $1000$ steps). The opaque curve is the mean of the results of $15$ experiments. The shadows are the bands of the confidence interval obtained by estimating the unbiased variance of the collective rewards. (e) The comparison between the mean collective rewards of all methods computed by removing the best and worst results of each method. The IA and baseline methods have a sharp initial raise but then slow down, whereas the EMuReL and SI have a relatively constant growth rate (EMuReL as the smoothest) and that EMuReL grows with a greater slop compared to SI.  (f) The comparison between the mean equality of all methods calculated by  using the Gini coefficient as $	Equality=1-\frac{\sum_{i=1}^{N}\sum_{j=1}^{N} |R^i-R^j|}{2N\sum_{i=1}^{N}R^i}$ \citep{hughes2018inequity}.}
		\label{fig.Reward_cleanup}
	\end{figure}
\end{center}

\begin{center}
	\begin{figure}[!h]
		\centering			
		\resizebox{\columnwidth}{!}{
			\extrarowsep=_3pt^3pt			
			\begin{tabu}to\linewidth{cccc}
				\raisebox{-0.5\height}{\includegraphics[width=0.5\columnwidth ]{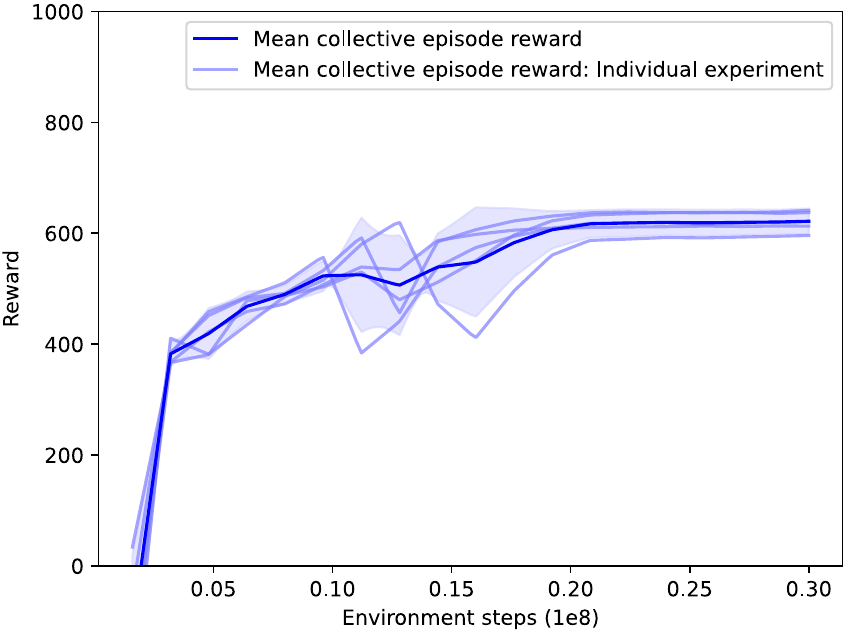}} &
				\raisebox{-0.5\height}{\includegraphics[width=0.5\columnwidth ]{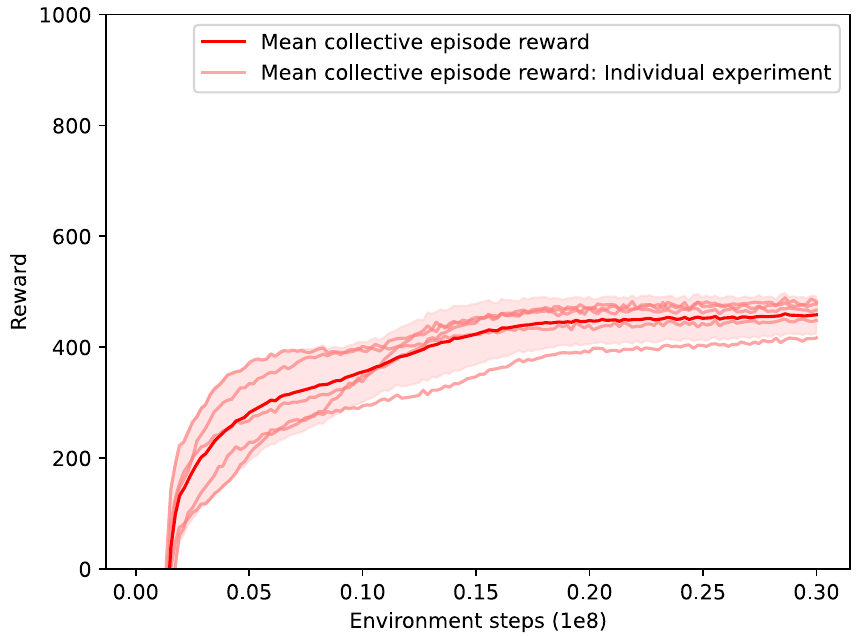}} & \raisebox{-0.5\height}{\includegraphics[width=0.5\columnwidth ]{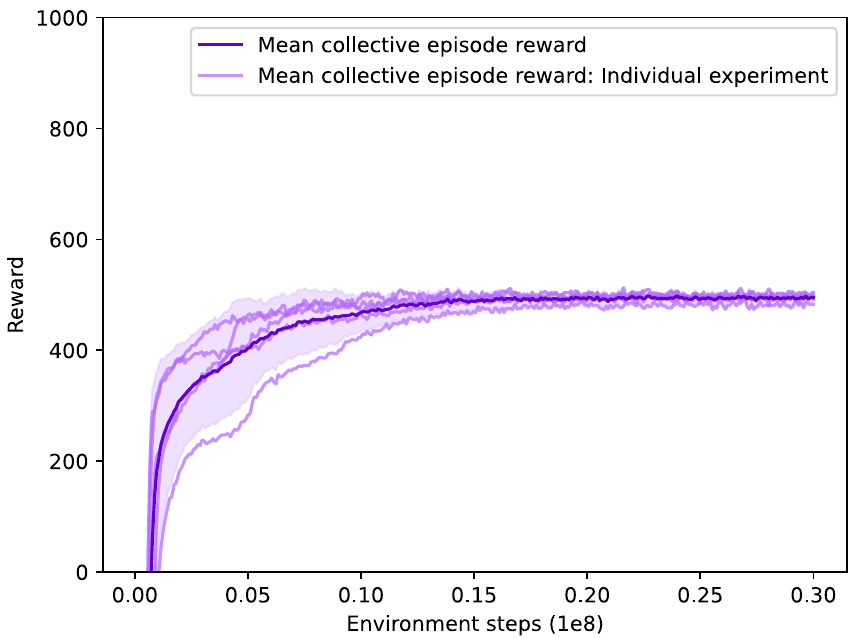}}	& \raisebox{-0.5\height}{\includegraphics[width=0.5\columnwidth ]{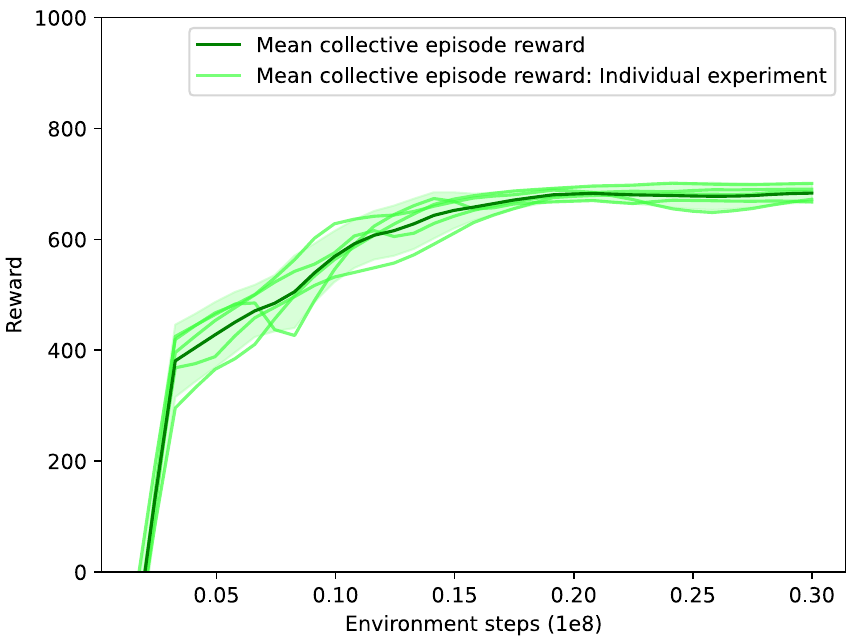}}
				\\
				\LARGE (a) Baseline & 
				\LARGE (b) SI & \LARGE (c) Advantageous IA & \LARGE (d)  Advantageous EMuReL  \\
				\raisebox{-0.5\height}{\includegraphics[width=0.5\columnwidth ]{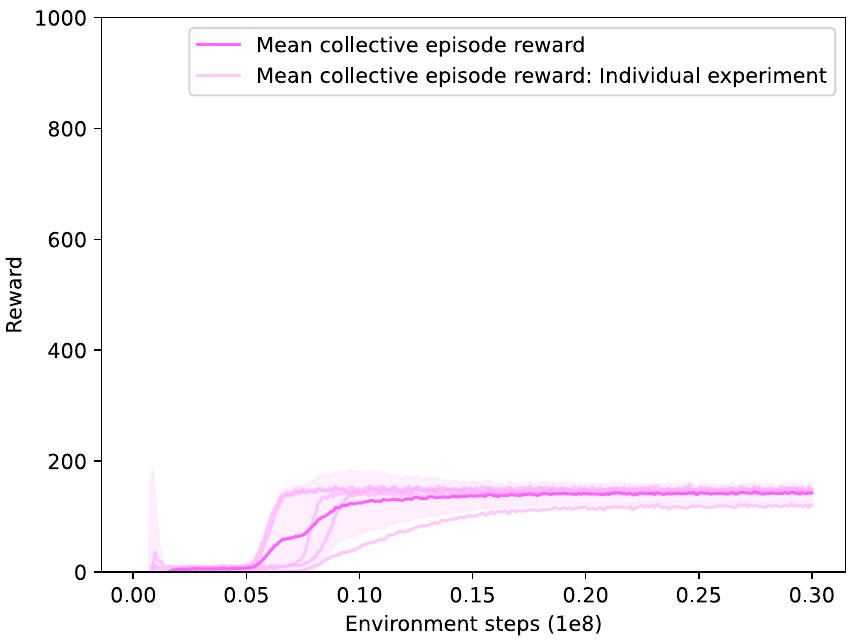}}	& \raisebox{-0.5\height}{\includegraphics[width=0.5\columnwidth ]{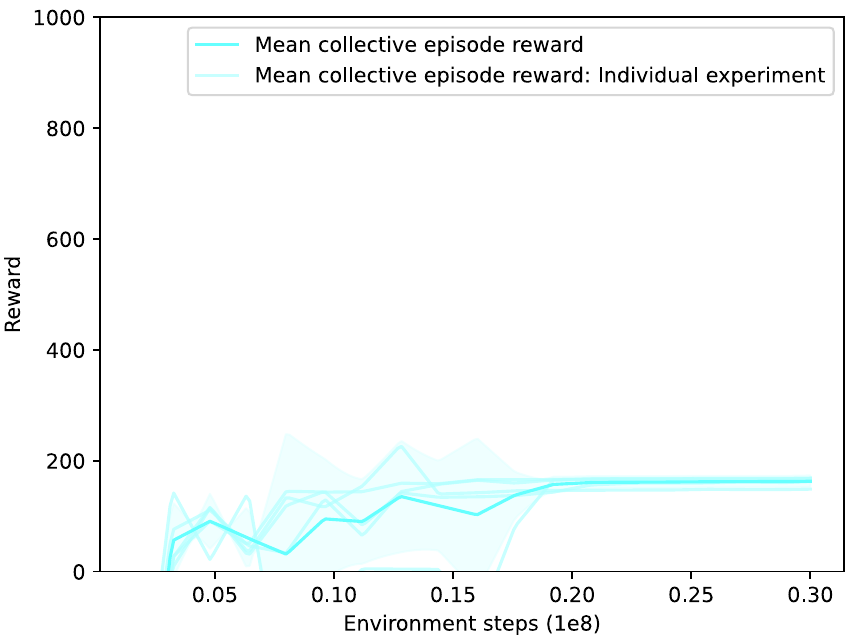}} & 	\multicolumn{2}{c}{\raisebox{-0.5\height}{\includegraphics[width=0.8\columnwidth ]{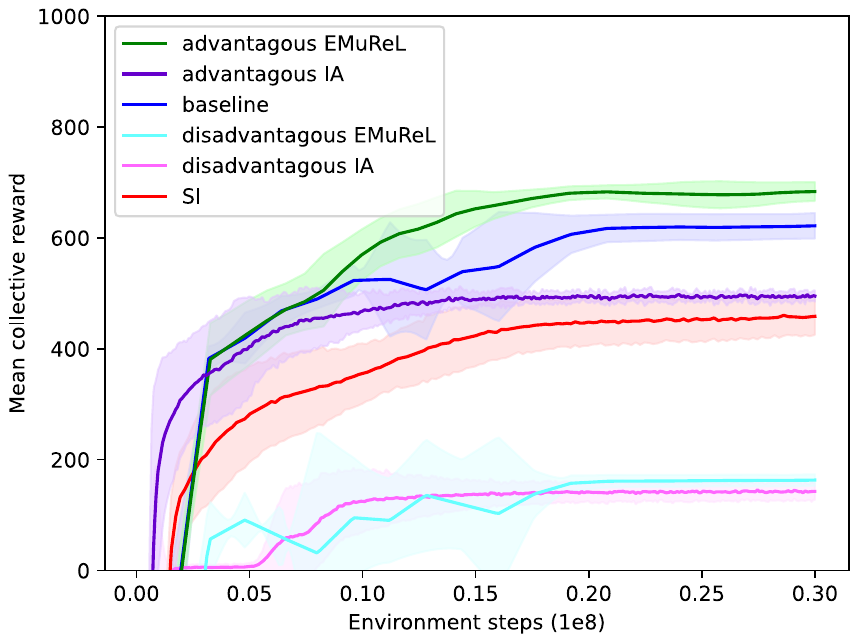}}} \\ 
				\\
				\LARGE (e) Disadvantageous IA & 
				\LARGE (f) Disadvantageous EMuReL & \multicolumn{2}{c}{\LARGE (g) Comparison} 
		\end{tabu}}
		\caption{\textbf{The results for the Harvest environment using A3C algorithm.} The same setup as that of Figure \ref{fig.Reward_cleanup} is used with the difference of using the A3C rather PPO algorithm and 4 instead of 5 agents. The opaque curve is the mean of the results of $5$ experiments. The curves of all methods become almost fixed after $\num{0.2e8}$ steps. The advantageous EMuReL method outperforms other methods after $\num{0.1e8}$ steps and has the overall better result.}
		\label{fig.Reward_harvest}
	\end{figure}
\end{center}

The total reward received by all agents is considered as a measure that explains how well the agents learned to cooperate \citep{hughes2018inequity}. According to numerical comparisons (Table \ref{Table.expriment_results}), when trained based on the EMuReL method, the agents can earn $53.6\%$ and $38.6\%$ more total rewards than when they were trained based on the IA model, $19.7\%$ and $44.2\%$ more than when they were trained based on SI method, and $36.9\%$ and $10.2\%$ more than when they were trained by using only the extrinsic rewards in the Cleanup and Harvest environments, respectively. Moreover, the EMuReL method persistently outperforms the other methods starting from step $2.15e8$ and has the minimum performance variance (subplot (e) in Figure \ref{fig.Reward_cleanup}). 
\begin{center}	
	\begin{table}[!h]
		\centering
		\caption{\textbf{Collected rewards in Cleanup and Harvest environments.} The collective reward obtained by agents in the last $\num{192000}$ and $\num{1600000}$ steps is averaged over $13$ and $5$ experiments in Cleanup and Harvest environments, respectively. The results of IA and EMuReL methods are reported for the advantageous-IA agents.}
		\label{Table.expriment_results}
		\resizebox{0.8\columnwidth}{!}{
			\extrarowsep=_3pt^3pt
			\begin{tabu}to\linewidth{|c|c|[2pt gray]c|c|c|c|}				
				\cline{3-6}
				\multicolumn{2}{c|[2pt gray]}{} & \multicolumn{4}{c|}{\textbf{Methods}} \\	
				\cline{3-6}
				\multicolumn{2}{c|[2pt gray]}{} & Baseline & IA & SI & EMuReL\\				
				\cline{2-6}				
				\tabucline[1.5pt gray]-
				\multirow{2}{*}{\textbf{Environment (Algorithm)}} &  Cleanup (PPO) & 484.4 &  431.7 &  554.2 & \textbf{663.2} \\
				\cline{2-6}
				& Harvest (A3C)  & 622.2 & 494.7 & 475.5 & \textbf{685.5} \\						
				\cline{1-6}
		\end{tabu}}
	\end{table}
\end{center}
All four methods perform almost identically in the Harvest environment using the PPO algorithm (See Appendix \ref{S.Appendix}). The same results are reported by \citet{heemskerk2020social} comparing the baseline, SI, and SCM methods when the PPO algorithm is applied. Since the results reported by \citet{jaques2019social,hughes2018inequity} for training the agents based on A3C algorithm achieved higher collective rewards compared to our PPO results, we also used the A3C instead of PPO algorithm to examine the performance of the EMuReL method. We compared the results with 4 agents in Figure \ref{fig.Reward_harvest}. The subplot (g) shows that the EMuReL method with advantageous-IA agents (called advantageous EMuReL method) can outperform other methods.
\section{Discussion}
\label{S.Discussion}
Researchers have developed several MARL methods to train cooperative agents in SSD problems\citep{canese2021multi, foerster2018counterfactual}. They introduced intrinsic rewards as a stimulus that is internally computed by agents to accelerate and improve their learning process \citep{singh2004intrinsically}. 
IA and SI are two state-of-the-art examples of reward reshaping methods, that are based on two social concepts: inequity aversion and social influence \citep{jaques2019social,hughes2018inequity}. 
We proposed the EMuReL method based on the social responsibility concept. To this end, we defined the environmental impact as a criterion to measure the role of each agent in reaching the current environment state and in turn, making the agents continuously measure the cooperativeness of their fellows. We incorporated these impacts into the reward function of the IA method. So, in the advantageous case, the more the agents play an impactful role in reaching the current environment state, the less each agent feels socially responsible for them and, as a result, the less it penalizes itself with negative internal rewards induced by the IA method. 

To compute the agents' impacts, inspired by the SCM method \citep{heemskerk2020social}, we proposed the EICM structure, that extends the single-agent curiosity setup of the ICM method \citep{pathak2017curiosity} to the MARL setting. The EICM structure utilizes the representation function $\phi$ to isolate the impact of each agent's action on the current environment state and to make the agents \qq{curious} about the behavior of others by assessing their environmental impacts. The EMuReL method achieved better results compared to the IA and SI methods in the Cleanup and  Harvest environments. According to the ablation results (see Supplementary material), the inclusion of each of the key ingredients of the EMuReL approach is necessary and contributes to the obtained performance. 

In the Cleanup environment, as \citep{hughes2018inequity} explains, the (advantageous) IA method encourages the agents to contribute and clean up the river. Cleaning up the waste causes producing more apples in the field and makes other agents more successful to collect more apples. So it reduces the negative rewards fed by the IA method. Here, incorporating the environmental impacts in the IA method improves the agents' policies. Depending on what action each agent chooses, its impact on the environment is remarkably different in this environment. If the agent chooses one of the movement actions without collecting apples, it creates the minimum change in the environment state, which is simply the change of its location. Such agents are subject to social responsibility by the other agents. If, however, the movement action is with collecting an apple, the field conditions additionally change. The most impactful agens are those who clean the river. Such an agent, in addition to changing the condition of the river on a large scale, changes the condition of the field by producing several apples depending on the growth rate of the apples. 

After completing the learning phase of the EMuReL method in the Cleanup environment, the behavior of the agents were investigated in the rendered videos for the polices learned in one of the last episodes. It was observed that one of the agents learned a circular movement pattern in the environment so that it travels the entire length of the river and then enters the apple field. On the way back to the river, it collects the apples. This agent also follows a specific pattern to clean up the river. It cleans the maximum possible width of the river whenever it applies the cleaning beam and cleans the whole width of the river every time it crosses the river. When returning to the river area, it chooses the shortest path according to the position of the apples in the field. The other agents maneuver in the apple field and collect apples. None of the agents in their behavior pattern use the fining beam; neither do they prevent each other from moving in the direction of the apples.

\subsection{Limitations and future work}
The proposed algorithm is distributed in the sense that no central agent is needed to learn the policies--each agent has its own MOA model. 
Nevertheless, the agents require access to all agents' actions. 
Relaxing this constraint to the case where the state in the absence of a certain agent is predicted using the actions of the local agents only is subject to future work. 
Other limitations include lacking comparisons to other more recent MARL algorithms and environments, long required training time, and considering only a discrete (rather than continuous) action space.



\begin{ack}
	We would like to thank Digital Research Alliance of Canada for providing computational resources that facilitated our experiments.
\end{ack}

\bibliographystyle{plainnat-reversed}
\bibliography{mybib}

\appendix
\renewcommand\thefigure{\thesection.\arabic{figure}}
\section{Appendix}
\label{S.Appendix}
\setcounter{figure}{0}  
\begin{center}
	\begin{figure}[!h]
		\centering
		\begin{tabular}{c@{ }c}
			\includegraphics[width=0.31\columnwidth]{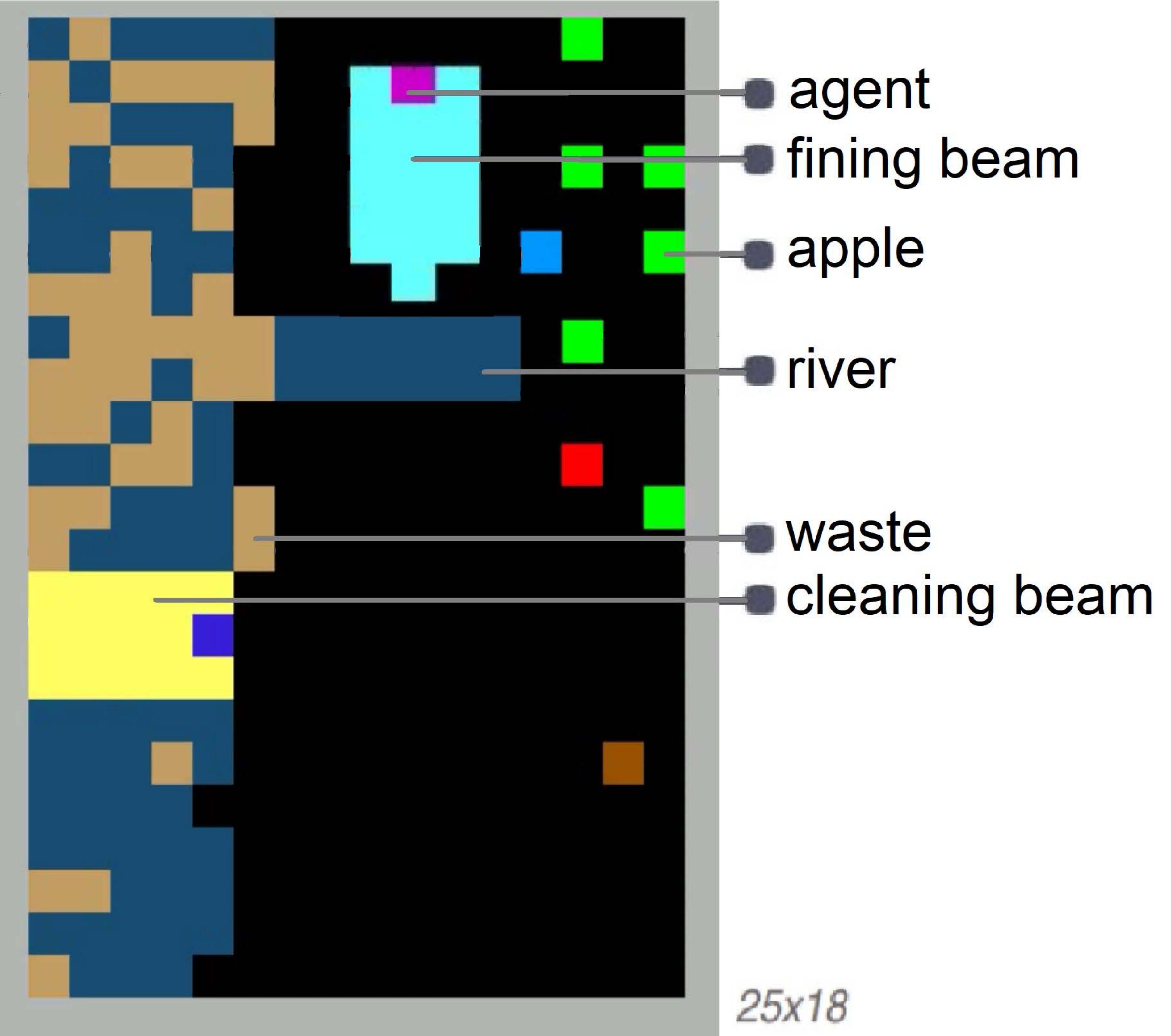}&		 
			\includegraphics[width=0.40\columnwidth]{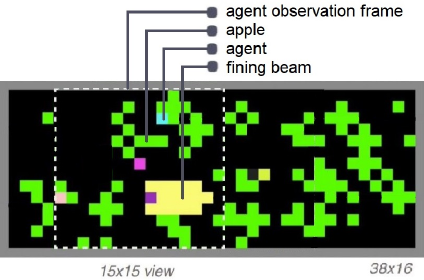} \\
			(a) & (b) 		
		\end{tabular}
		\caption{\textbf{The SSD environments.} (a) Cleanup and (b) Harvest games.} 
		\label{fig.1_cleanup_harvest}
	\end{figure}
\end{center}
\begin{center}
	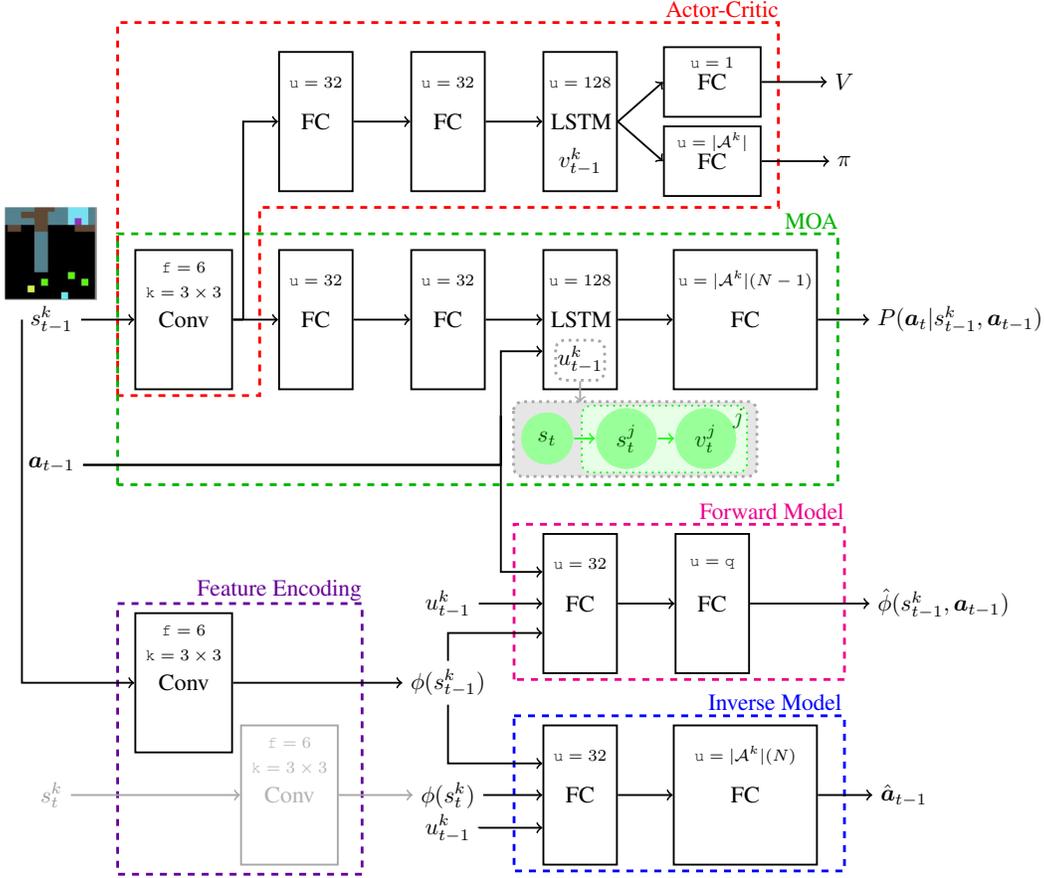
\begin{figure}[!h]
		\centering
		\resizebox{\columnwidth}{!}{
			\begin{tikzpicture}
			\node[inner sep=0pt] (image) at (0,1)
			{\includegraphics[width=.1\textwidth]{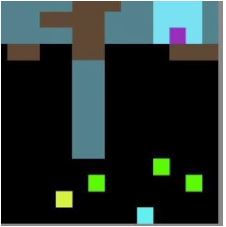}};
			\node (Input1) at (0,0) [] {$s^k_{t-1}$};
			\node (Input2) at (0,-2.2) [] {$\bm{a}_{t-1}$};
			\node (Conv) at (2,0) [layers, rectangle, text width=3.5em] {Conv};
			\node (Convk) at (2,0.4) [] {\scriptsize $\texttt{k}=3\times3$};
			\node (Convf) at (2,0.8) [] {\scriptsize $\texttt{f}=6$};
			\node (FC1)  at (4,0) [layers, rectangle] {FC};
			\node (FC1u) at (4,0.6) [] {\scriptsize $\texttt{u}=32$};
			\node (FC2)  at (6,0) [layers, rectangle] {FC};
			\node (FC2u) at (6,0.6) [] {\scriptsize $\texttt{u}=32$};
			\node (LSTM) at (8,0) [layers, rectangle] {LSTM};
			\node (LSTMu) at (8,0.6) [] {\scriptsize $\texttt{u}=128$};
			\node (ukt) at (8,-0.6) [] {$u^k_{t-1}$};
			\node (FC3)  at (10.5,0) [layers, rectangle, text width=5.5em] {FC};
			\node (FC3u) at (10.5,0.6) [font=\scriptsize] { $\texttt{u}=|\mathcal{A}^k|(N-1)$};
			\node (Output) at (13.75,0) [] {$P(\bm{a}_{t}|s^k_{t-1},\bm{a}_{t-1})$ };
			\draw [->, thick] (Input1.east) -- (Conv.west);  
			\draw [->, thick] (Conv.east) --  (FC1.west);
			\draw [->, thick] (FC1.east) --  (FC2.west);
			\draw [-, thick] (Input2) -| (6.8,-1.45) ;
			\draw [->, thick] (6.8,-1.45) |- (LSTM.220) ;
			\draw [->, thick] (FC2.east) --  (LSTM.west);    
			\draw [->, thick] (LSTM.east) --  (FC3.west);    
			\draw [->, thick] (FC3.east) -- (Output.west);        
			\node (st) at (7.5,-1.8) [states,text=black!70!green, circle] {$s_t$};   
			\node (sjt) at (8.7,-1.8) [states,text=black!70!green, circle] {$s^j_t$};    
			\node (vjt) at (9.9,-1.8) [states,text=black!70!green, circle] {$v^j_t$};    
			\draw [->, thick, green!90] (st.east) -- (sjt.west);
			\draw [->, thick, green!90] (sjt.east) -- (vjt.west);  			        
			\node (MOA) at (11.5,1.5) [text=black!30!green] {MOA}; 
			
			\node (acFC1)  at (4,3) [layers, rectangle] {FC};
			\node (acFC1u) at (4,3.6) [] {\scriptsize $\texttt{u}=32$};
			\node (acFC2)  at (6,3) [layers, rectangle] {FC};
			\node (acFC2u) at (6,3.6) [] {\scriptsize $\texttt{u}=32$};
			\node (acLSTM) at (8,3) [layers, rectangle] {LSTM};
			\node (acLSTMu) at (8,3.6) [] {\scriptsize $\texttt{u}=128$};
			\node (acvkt) at (8,2.4) [] {$v^k_{t-1}$};
			\node (acFC3V)  at (10,3.6) [layers, rectangle,minimum height=3em, text width=3.5em] {FC};
			\node (acFC3Pi)  at (10,2.4) [layers, rectangle,minimum height=3em, text width=3.5em] {FC};
			\node (acFC3Vu) at (10,3.9) [font=\scriptsize] { $\texttt{u}=1$};
			\node (acFC3Piu) at (10,2.7) [font=\scriptsize] { $\texttt{u}=|\mathcal{A}^k|$};
			\node (acOutputV) at (12,3.6) [] {$V$};
			\node (acOutputPi) at (12,2.4) [] {$\pi$};  
			\draw [-, thick] (Conv.east) --  (2.9,0);
			\draw [->, thick] (2.9,0) |-  (acFC1.west);
			\draw [->, thick] (acFC1.east) --  (acFC2.west);
			\draw [->, thick] (acFC2.east) --  (acLSTM.west);    
			\draw [->, thick] (acLSTM.east) --  (acFC3V.west);  
			\draw [->, thick] (acLSTM.east) --  (acFC3Pi.west);  
			\draw [->, thick] (acFC3V.east) -- (acOutputV.west);
			\draw [->, thick] (acFC3Pi.east) -- (acOutputPi.west);
			\node (ActorCritic) at (10.15,4.7) [text=red] {Actor-Critic};
			
			\node (FAConv1) at (2,-5.5) [layers, rectangle, text width=3.5em] {Conv};
			\node (FAConvk1) at (2,-5.1) [] {\scriptsize $\texttt{k}=3\times3$};
			\node (FAConvf1) at (2,-4.7) [] {\scriptsize $\texttt{f}=6$};
			\draw [->, thick] (Input1.west) |- (FAConv1.west);
			\node (PhiPrevSInput) at (6,-5.5) [] {$\phi(s^k_{t-1})$};
			\draw [->, thick] (FAConv1.east) |- (PhiPrevSInput.west);
			\node (SA2Input) at (0,-7.2) [text=gray!80] {$s^k_{t}$};
			\node (FAConv2) at (3.6,-7.2) [layers, rectangle, draw=gray!70, text=gray!60, text width=3.5em] {Conv};
			\node (FAConvk2) at (3.6,-6.8) [text=gray!60] {\scriptsize $\texttt{k}=3\times3$};
			\node (FAConvf2) at (3.6,-6.4) [text=gray!60] {\scriptsize $\texttt{f}=6$};
			\draw [->, thick, draw=gray!70] (SA2Input.east) -- (FAConv2.west);
			\node (PhiSInput) at (6,-7.2) [] {$\phi(s^k_{t})$};
			\draw [->, thick, draw=gray!70] (FAConv2.east) |- (PhiSInput.west);
			\node (FeatureEncoding) at (3.45,-4.1) [text=blue!60!red] {Feature Encoding};
			
			\node (uktInputF) at (6,-4.3) [] {$u^k_{t-1}$};
			\node (FFC1)  at (8,-4.3) [layers, rectangle] {FC};
			\node (FFC1u) at (8,-3.7) [] {\scriptsize $\texttt{u}=32$};
			\node (FFC2)  at (10,-4.3) [layers, rectangle] {FC};
			\node (FFC2u) at (10,-3.7) [] {\scriptsize $\texttt{u}=\texttt{q}$};
			\node (FOutput) at (13.5,-4.3) [] {$\hat{\phi}(s^k_{t-1},\bm{a}_{t-1})$};
			\draw [-, thick] (Input2) -| (6.8,-1.45) ;
			\draw [->, thick] (6.8,-1.45) |- (FFC1.140);
			\draw [->, thick] (uktInputF.east) |- (FFC1.180);
			\draw [->, thick] (PhiPrevSInput.north) |- (FFC1.218);
			\draw [->, thick] (FFC1.east) |- (FFC2.west);
			\draw [->, thick] (FFC2.east) |- (FOutput.west);
			\node (ForwardModel) at (10.9,-2.9) [text=magenta] {Forward Model};
			
			\node (uktInputI) at (6,-7.7) [] {$u^k_{t-1}$};
			\node (IFC1)  at (8,-7.2) [layers, rectangle] {FC};
			\node (IFC1u) at (8,-6.6) [] {\scriptsize $\texttt{u}=32$};
			\node (IFC2)  at (10.5,-7.2) [layers, rectangle, text width=5.5em] {FC};
			\node (IFC2u) at (10.5,-6.6) [] {\scriptsize $\texttt{u}=|\mathcal{A}^k|(N)$};
			\node (IOutput) at (12.9,-7.2) [] {$\hat{\bm{a}}_{t-1}$};
			\draw [->, thick] (PhiPrevSInput.south) |- (IFC1.140);
			\draw [->, thick] (PhiSInput.east) |- (IFC1.180);
			\draw [->, thick] (uktInputI.east) |- (IFC1.221);
			\draw [->, thick] (IFC1.east) |- (IFC2.west);
			\draw [->, thick] (IFC2.east) |- (IOutput.west);
			\node (InverseModel) at (10.95,-5.8) [text=blue] {Inverse Model};
			
			\begin{pgfonlayer}{background}  
			
			\path (st.west |- st.north)+(-0.1,0.15) node (c) {};
			\path (vjt.south -| vjt.east)+(+0.3,-0.1) node (d) {};
			\path (sjt.west |- sjt.north)+(-0.2,0.02) node (e1) {};
			\path (vjt.south -| vjt.east)+(+0.15,-0.04) node (e2) {};
			\path[fill=gray!20,rounded corners, draw=gray!70,very thick, dotted]
			(c) rectangle (d);
			\path[fill=green!10,rounded corners, draw=green,thick, dotted]
			(e1) rectangle (e2);	
			
			\path[draw=red, dashed, very thick] (1,-1.15)-- (1,4.5)--(11,4.5)-- (11,1.7)--(3.15,1.7)--(3.15,-1.15)--(1,-1.15);
			
			\path[draw=black!30!green, dashed, very thick]
			(1,-2.5) rectangle (11.9,1.3); 
			
			\path[draw=blue!60!red, dashed, very thick] (1,-4.3)-- (4.7,-4.3)--(4.7,-8.4)-- (1,-8.4)--(1,-4.3); 
			
			\path[draw=magenta, dashed, very thick] (7,-5.45)-- (7,-3.1)--(12,-3.1)-- (12,-5.45)--(7,-5.45); 
			
			\path[draw=blue, dashed, very thick] (7,-6)-- (12,-6)-- (12,-8.35)--(7,-8.35)--(7,-6); 
			
			\end{pgfonlayer}
			\begin{pgfonlayer}{foreground} 
			\path (LSTM.west |- LSTM.north)+(+0.2,-1.38) node (ukt1) {};
			\path (LSTM.south -| LSTM.east)+(-0.2,+0.15) node (ukt2) {};
			\path[rounded corners, draw=gray!70, very thick, dotted]
			(ukt1) rectangle (ukt2);   
			\draw [->, draw=gray!70, thick] (ukt.south) -- (8,-1.25); 
			\node (j) at (10.4,-1.5) [text=black!50!green] {$j$};
			\end{pgfonlayer}		
			\end{tikzpicture}}
		\caption{\textbf{The EICM network structure of the EMuReL method.} The EICM of each agent $k$ has two inputs: the previous local state $s^k_{t-1}$, for example an image (represented on its top), and the previous joint action $\bm{a}_{t-1}$. The EICM includes five distinctive networks: \emph{(i)} the actor-critic structure that learns the policy and value function, \emph{(ii)} the MOA network, \emph{(iii)} the feature extraction network that contains a convolutional layer (Conv) as an encoder to represent the local state $s^k_{t-1}$ as \texttt{q} features, \emph{(iv)} the forward model that learns $\hat\phi$, and \emph{(v)} the inverse model that predicts the applied actions. The Conv layer represented by a transparent gray rectangle inside the purple dashed rectangle indicates that the feature encodings of the local states $s^k_{t-1}$ and $s^k_t$ are done by using the same Conv layer, but at two consequence time steps $t-1$ and $t$. The parameters \texttt{k} and \texttt{f} in Conv present the convolution kernel size and filters, respectively. Parameter \texttt{u} is the number of neurons in FC layers. Each agent $j$ applies an action $a^{j}_{t}$ based on its local state $s^j_{t}$ and internal LSTM state $v^j_{t}$ of its actor-critic structure which serves as a memory of the previous states. Hence, to predict the action $a^{j}_{t}$ of each agent $j$, agent $k$'s MOA implicitly models the local state $s^j_t$ and internal LSTM state $v^j_t$ of all agents, which are implicitly captured in its internal LTSM state $u^k_{t-1}$ (the green nodes in the gray rectangle). As the aggregation of the local states $s^j_t$ form the global state $s_t$, the internal LSTM state $u^k_{t-1}$ can provide an estimation of the global state. The actor-critic and MOA structures are taken from \citet{jaques2019social}. The forward and inverse model are based on the structures presented by \citet{heemskerk2020social}.} 
		\label{fig.EICM}
	\end{figure}
\end{center}
\begin{center}
	\begin{figure}[!h]
		\centering			
		\resizebox{\columnwidth}{!}{
			\extrarowsep=_3pt^3pt			
			\begin{tabu}to\linewidth{c@{\hspace{-0.5cm}}c@{\hspace{-0.5cm}}c}
				\raisebox{-0.5\height}{\includegraphics[width=0.5\columnwidth ]{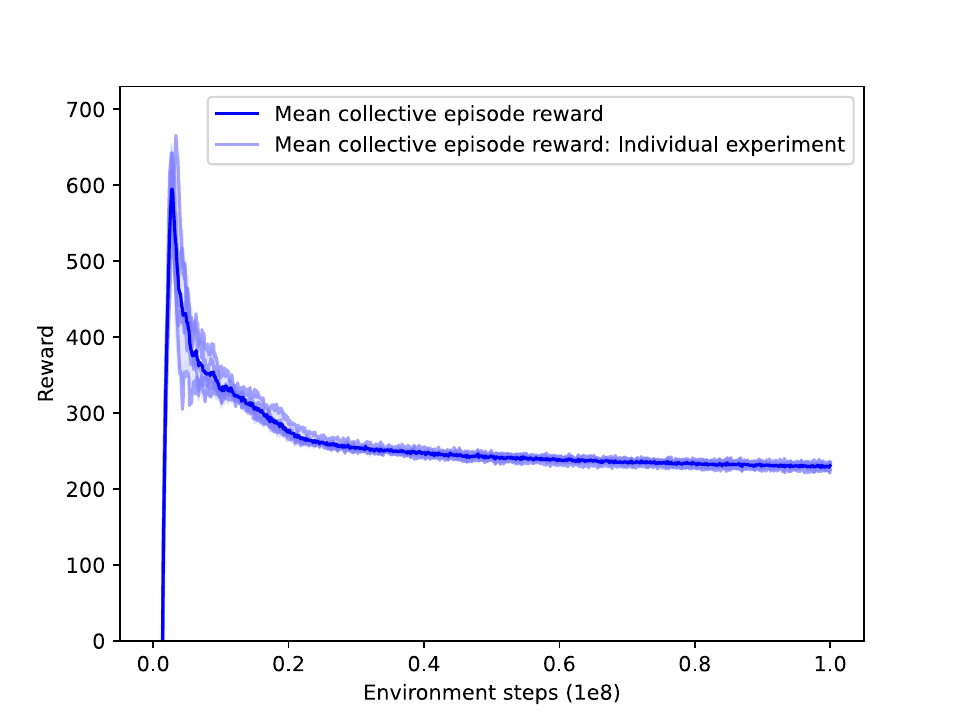}} &
				\raisebox{-0.5\height}{\includegraphics[width=0.5\columnwidth ]{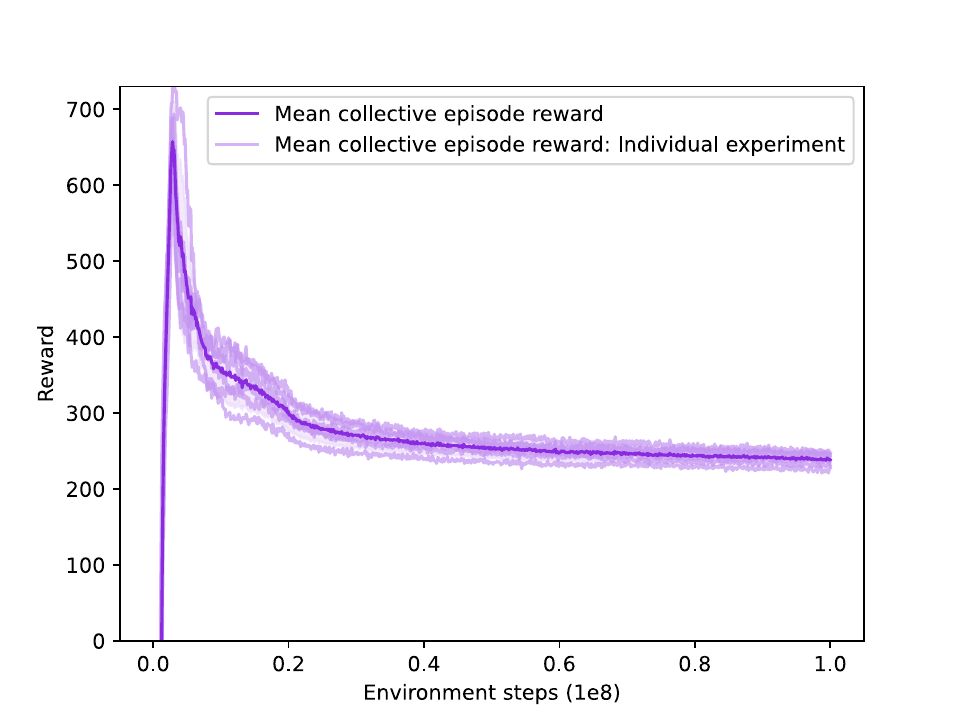}} &  
				\multirow{3}{*}{\raisebox{1.5\height}{\includegraphics[width=\columnwidth ]{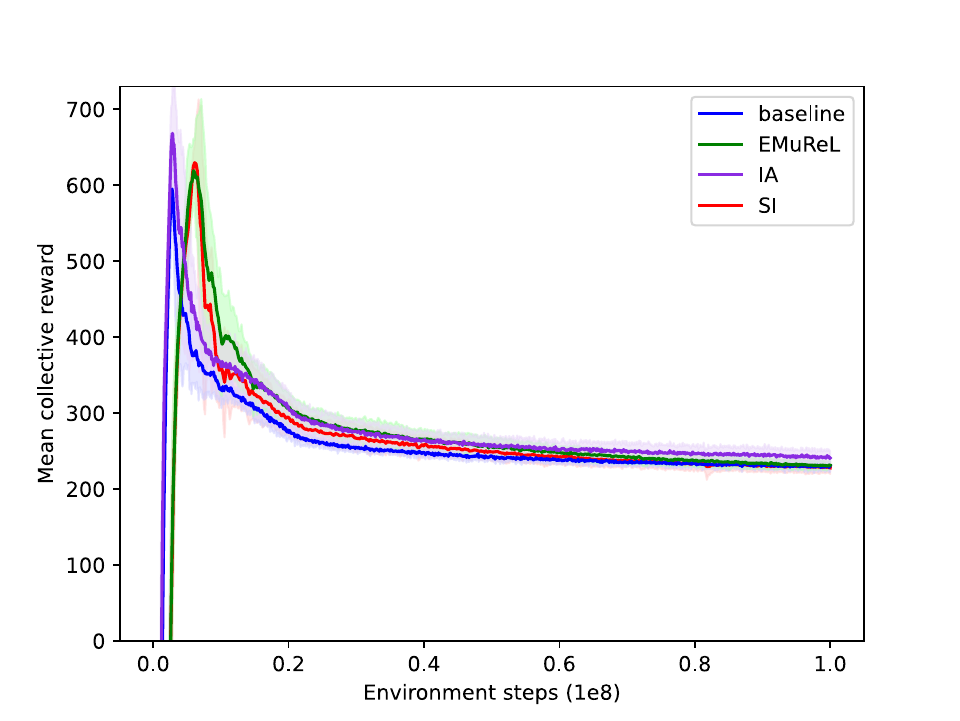}}}
				\\
				\LARGE (a) Baseline & 
				\LARGE (b) IA &  \\				
				\raisebox{-0.5\height}{\includegraphics[width=0.5\columnwidth ]{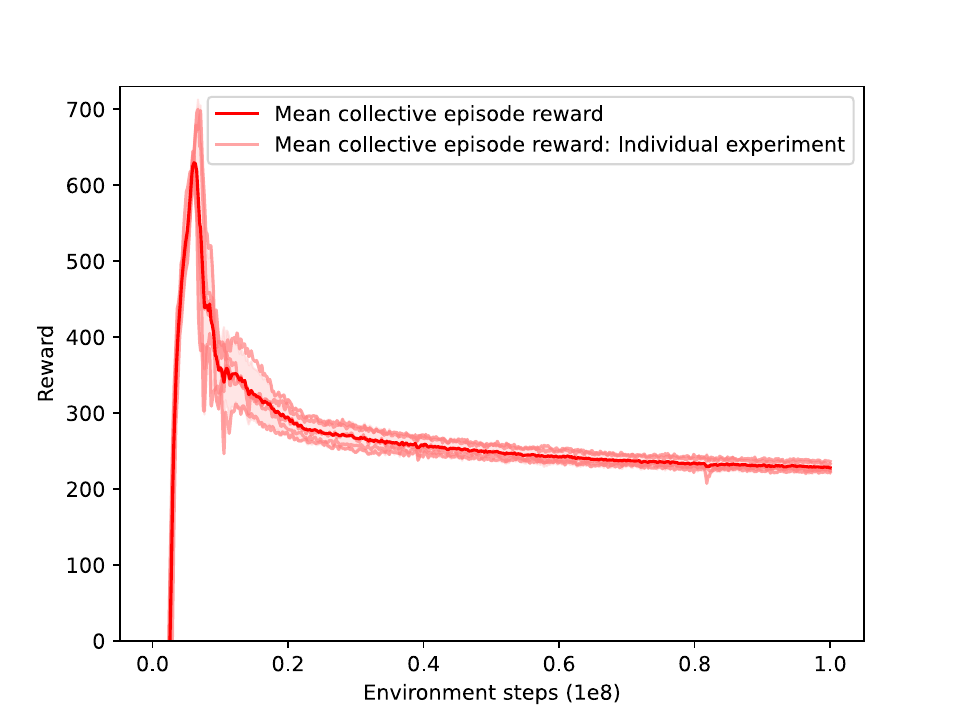}}	& \raisebox{-0.5\height}{\includegraphics[width=0.5\columnwidth ]{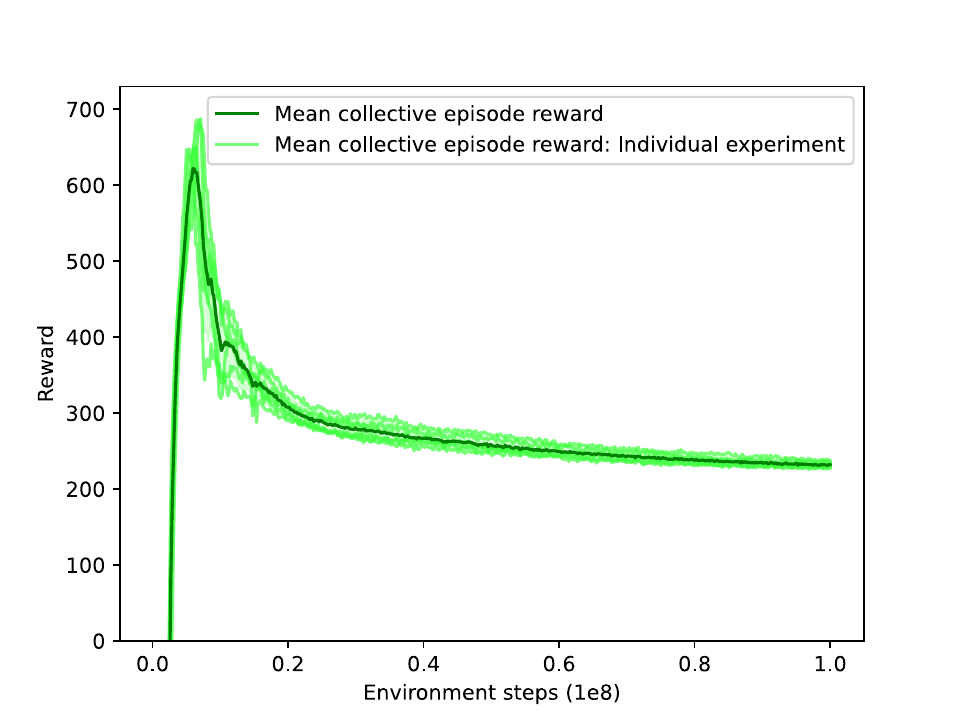}} &  \\
				\LARGE (c) SI	 & \LARGE (d)  EMuReL  & \LARGE (e) Comparison	 
		\end{tabu}}
		\caption{\textbf{The results for the Harvest environment using PPO algorithm.} The same setup as that of Figure \ref{fig.Reward_cleanup} is used. The opaque curve is the mean of the results of $5$ experiments. The final more-stable part of the results of all methods is almost identical in this experiment.} 
		\label{fig.Reward_harvest_comparison_PPO}
	\end{figure}
\end{center}
\end{document}